\documentclass{article}
\usepackage{ijcai24}
\usepackage{multirow}
\usepackage{times}
\usepackage{soul}
\usepackage{url}
\usepackage[utf8]{inputenc}
\usepackage[small]{caption}
\usepackage{graphicx}
\usepackage{amsmath}
\usepackage{amsthm}
\usepackage{booktabs}
\usepackage{algorithm}
\usepackage{algorithmic}
\usepackage[switch]{lineno}
\usepackage{xspace}
\usepackage{overpic}
\usepackage{amssymb}
\usepackage[pagebackref,breaklinks,colorlinks,citecolor=black]{hyperref}
\newcommand\blfootnote[1]{%
\begingroup
\renewcommand\thefootnote{}\footnote{#1}%
\addtocounter{footnote}{-1}%
\endgroup
}
\newcommand{\sysname}{{G2LTraj}\xspace}
\newcommand\figref[1]{Fig.~\ref{#1}}

\newcommand{\eg}{\emph{e.g.},\xspace}
\newcommand{\ie}{\emph{i.e.},\xspace}

\newcommand\tabref[1]{Table~\ref{#1}}

\newcommand\secref[1]{Sec.~\ref{#1}}
\newcommand\equref[1]{Eq.~(\ref{#1})}

\title{G2LTraj: A Global-to-Local Generation Approach for Trajectory Prediction}

\author{
Zhanwei Zhang$^{1*}$
\and
Zishuo Hua$^{2*}$\and
Minghao Chen$^{3}$\and
Wei Lu$^{5}$\and
Binbin Lin$^{2,4}$\and \\
Deng Cai$^{1}$\And
Wenxiao Wang$^{2\dag}$\\
\affiliations
$^1$State Key Lab of CAD\&CG, Zhejiang University\\
$^2$School of Software Technology, Zhejiang University\\
$^3$Hangzhou Dianzi University
$^4$Fullong Inc.\\
$^5$Ningbo Beilun Third Container Terminal Co., Ltd\\
\emails
\{zhanweizhang, 22251168, binbinlin, wenxiaowang\}@zju.edu.cn, \\
\{minghaochen01, dengcai\}@gmail.com,
luwei1@nbport.com.cn
}

\begin{document}

\maketitle

\begin{abstract}
Predicting future trajectories of traffic agents accurately holds substantial importance in various applications such as autonomous driving. 
Previous methods commonly infer all future steps of an agent either recursively or simultaneously. 
However, the recursive strategy suffers from the accumulated error,
while the simultaneous strategy overlooks the constraints among future steps, resulting in kinematically infeasible predictions.
To address these issues, in this paper, we propose \sysname, a plug-and-play global-to-local generation approach for trajectory prediction. 
Specifically, we generate a series of global key steps that uniformly cover the entire future time range.
Subsequently, the local intermediate steps between the adjacent key steps are recursively filled in.
In this way, we prevent the accumulated error from propagating beyond the adjacent key steps.
Moreover, to boost the kinematical feasibility, we not only introduce the spatial constraints among key steps but also strengthen the temporal constraints among the intermediate steps.
Finally, to ensure the optimal granularity of key steps, we design a selectable granularity strategy that caters to each predicted trajectory. 
Our \sysname significantly improves the performance of seven existing trajectory predictors across the ETH, UCY and nuScenes datasets.
Experimental results demonstrate its effectiveness.
Code will be available at \url{https://github.com/Zhanwei-Z/G2LTraj}.
\blfootnote{$^*$Equal contribution. $^\dag$Corresponding author}
\end{abstract}
\section{Introduction}
Accurate trajectory prediction is crucial in numerous applications like self-driving, enabling autonomous vehicles to make safe and reliable decisions. 
This task involves predicting the future trajectories of moving agents (\eg vehicles and pedestrians) by learning their interactions within the scenarios.
Recently, deep learning methods have come to dominate the field of trajectory prediction.
\begin{figure}[!t]
\centering
\begin{overpic}[width=\linewidth]{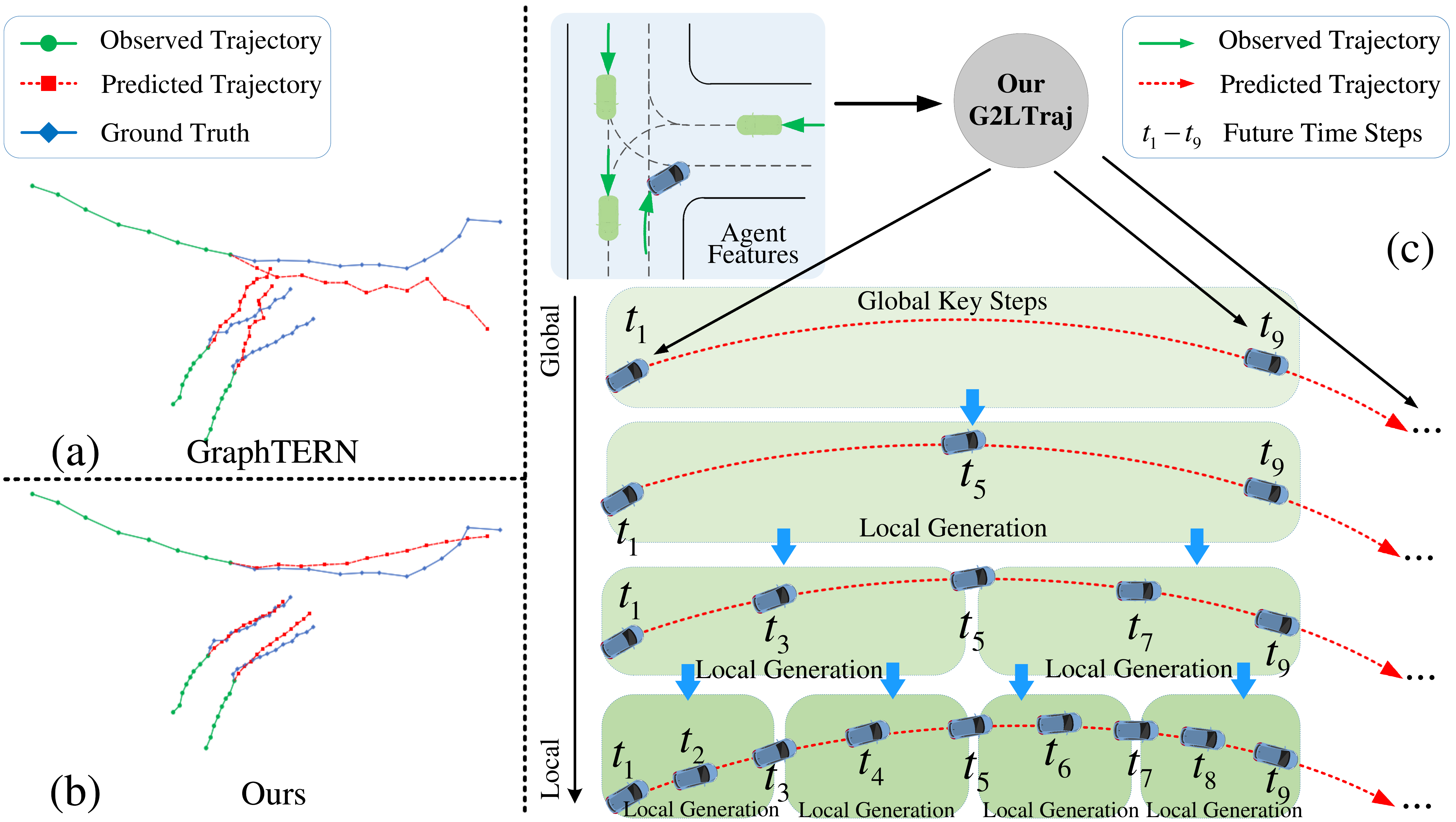}
\end{overpic}
\caption{
(a) State-of-the-art simultaneous approaches like GraphTERN {\protect\cite{bae2023set}} do not model the constraints among future steps well, resulting in kinematically infeasible predictions (\ie the fluctuation in the predicted trajectories), as evaluated on the UCY \protect\cite{lerner2007crowds} dataset.
(b) Our \sysname incorporates spatial-temporal constraints to generate consistent trajectories.
(c) Overview of our \sysname for trajectory prediction in a global-to-local generation process.
For instance, 
initially, we simultaneously generate the global key steps $t_1$, $t_9$ and so on. Subsequently, $t_5$ is generated by utilizing $t_1$ and $t_9$.
This iterative process continues until all future steps are generated.
}
\label{overview}
\end{figure}
These methods typically predict all future steps of an agent either simultaneously \cite{shi2021sgcn,shi2022motion,10192373,girgis2022latent,li2021spatial,bae2023eigentrajectory} or recursively \cite{yuan2021agentformer,chen2021personalized,gu2022stochastic,lee2022muse,zhao2021you}.
Specifically, 
the simultaneous methods assume that the future steps of an agent are independent of each other and commonly generate all future steps at once.
One branch of these methods \cite{aydemir2023adapt,bae2023set,gu2021densetnt,mangalam2021goals}
leverages trajectory anchors or endpoints to generate the intermediate steps concurrently.
While these methods have demonstrated promising performance enhancements, the constraints among future steps are ignored, potentially leading to kinematically infeasible predictions, as shown in \figref{overview}(a).
On the other hand, to capture the temporal constraints among future steps, the recursive strategy utilizes previously predicted time steps to forecast the subsequent ones.
However, this strategy suffers from 
the accumulated error throughout the recursive process \cite{bae2023set}, with the error propagating from the initial step to the final step.
To balance the accumulated error and the constraints among future steps, \cite{jia2023towards} simultaneously generates all future steps and utilizes a recursive network for trajectory refinement. 
Nonetheless, during refinement, the accumulated error still propagates from the initial step to the final step.

To tackle the above issues, in this paper, we propose G2LTraj, a plug-and-play global-to-local generation approach for trajectory prediction.
Specifically, we generate a series of global key steps that uniformly span the entire future time range, establishing the trajectory outline.
We treat any two adjacent key steps as the head and the tail steps of a local section.
By partitioning the entire trajectory into multiple sections, the accumulated error would not propagate beyond each section.
Subsequently, the intermediate steps within each section are recursively generated to complete the whole trajectory.
We illustrate the overview of our \sysname in \figref{overview}(c) intuitively.
During the global-to-local generation process, we identify three specific challenges requiring attention.
\textit{{\textbf{First}}}, during the global key step generation process,
since all key steps are generated at once, there is a lack of constraints among them.
To this end, we introduce the spatial constraints among key steps by regulating the positional difference.
This enables \sysname to generate consistent trajectories that adhere to kinematics, as shown in \figref{overview}(b).
\textit{{\textbf{Second}}},
during the local recursive generation process,
if merely relying on the head and tail steps to generate the intermediate step, 
the model's access to past kinematic information is limited, resulting in fragile temporal constraints with past trajectories.
This fragility possibly leads to plausible but kinematically infeasible intermediate steps.
To maximize the temporal constraints, 
we integrate the agent features along with the position information of the head and tail steps when generating the intermediate step.
By doing so, \sysname can enhance its understanding of local motion patterns.
\textit{{\textbf{Third}}}, 
trajectories from different agents can exhibit varying kinematics characteristics like distinct velocities.
Thus, using a fixed granularity (\ie the time interval between two adjacent key steps) might not be suitable for each trajectory, which aligns with previous findings \cite{bae2023set}.
Distinguished from manually adjusting the granularity for all trajectories \cite{bae2023set}, we design a selectable granularity strategy tailored for each trajectory.
Specifically,
we generate key steps at a fine granularity and then downsample these steps to form multiple groups of coarse-grained key steps.
Next, by learning the confidence scores of these groups for each trajectory, we select the optimal group for them.
Our contributions can be summarized as follows:
\begin{itemize}
\item we propose a global-to-local generation approach for trajectory prediction that mitigates the accumulated error and introduces the constraints among future steps.
\item  To enhance the kinematical feasibility, we propose the spatial constraints among key steps and reinforce the temporal constraints among the intermediate steps.
\item To ensure the optimal granularity of key steps for each trajectory, we devise a selectable granularity strategy. 
\item Our \sysname demonstrates notable performance gains over seven existing trajectory predictors when evaluated on ETH \cite{pellegrini2009you}, UCY \cite{lerner2007crowds} and nuScenes \cite{caesar2020nuscenes}.
Extensive experiment results validate its effectiveness.
\end{itemize}
\begin{figure*}
\centering
\begin{overpic}[width=\linewidth]{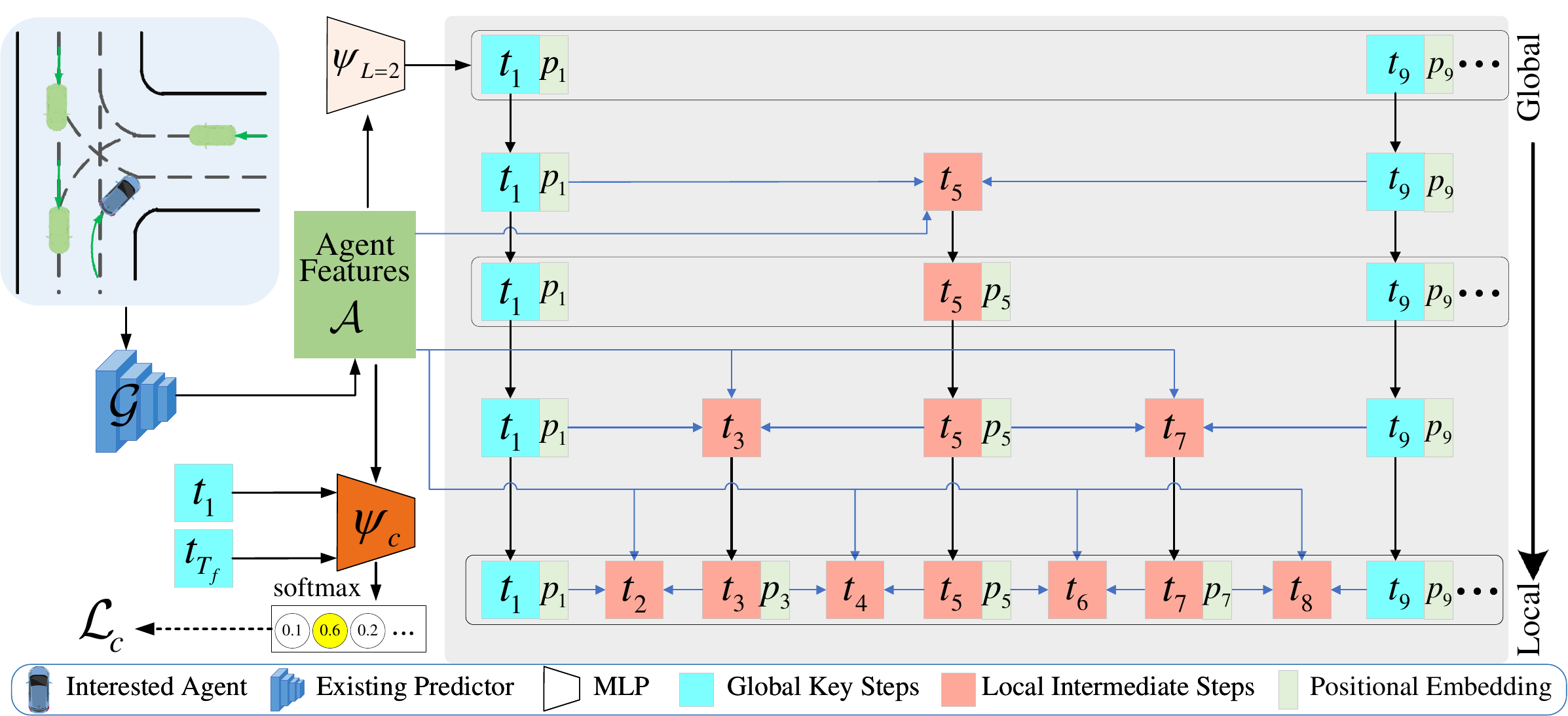}
\end{overpic}
\caption{The overview framework of our \sysname. 
We introduce a global-to-local generation approach for trajectory prediction that attenuates the accumulated error and proposes constraints among future steps.
In this illustration, we exemplify the global-to-local generation process at a granularity of 8.
we simultaneously generate the globakey steps $t_1$, $t_9$ and so on. Subsequently, $t_5$ is generated by utilizing $t_1$ and $t_9$.
This local recursion process continues until all future steps are generated.}
\label{framework}
\end{figure*}

\section{Related Work}
Accurate trajectory prediction plays a vital role in various applications such as self-driving. 
Recent state-of-the-art methods primarily rely on neural networks to forecast all future steps of an agent, either simultaneously or recursively.

{\setlength{\parindent}{0cm}\textbf{Simultaneous Trajectory Predictors}} assume that the future steps of an agent are independent of each other and typically generate future steps concurrently \cite{10192373,shi2021sgcn,shi2022motion,girgis2022latent}.
\cite{shi2022social} utilizes tree paths based on the observed trajectories to predict future trajectories in parallel. 
\cite{shi2021sgcn} and \cite{li2021spatial} achieve simultaneous trajectory prediction by capturing both spatial and temporal clues.
\cite{zhou2023query} endeavors to achieve faster inference while maintaining effectiveness.
Another branch \cite{bae2023set,aydemir2023adapt,gu2021densetnt} initially learns the potential endpoints or anchor points of local trajectories and subsequently generates the intermediate steps simultaneously.
For example, \cite{aydemir2023adapt} employs the predicted endpoints to jointly predict the trajectories of all agents in the scene.
\cite{ye2022dcms} considers both spatial and temporal perturbations when predicting the endpoint of future trajectories. 
However, these methods commonly neglect the constraints among future steps, possibly resulting in predictions that are kinematically infeasible \cite{jia2023towards}.

{\setlength{\parindent}{0cm}\textbf{Recursive Trajectory Predictors}} utilizes previously predicted time steps to generate the subsequent ones \cite{gu2022stochastic,lee2022muse,yuan2021agentformer,chen2021personalized,zhao2021you,salzmann2020trajectron++,xu2020cf}.
\cite{marchetti2022smemo} and \cite{zhao2021you} employ supplementary storage space to generate future trajectories sequentially. 
\cite{navarro2022social} proposes guiding long-term time-series trajectory prediction in complex environments. 
\cite{gu2022stochastic} progressively discards area indeterminacy until reaching the desired trajectory.
\cite{lee2022muse} leverages RNN-based Conditional
Variational AutoEncoder (CVAE) to obtain the full trajectories.
Moreover, \cite{jia2023towards} exploits a recursive network for trajectory refinement while simultaneously generating all future steps.
Despite capturing temporal constraints, these methods encounter the accumulated error throughout the recursive process, as the error propagates from the initial step to the final step.

To this end, we propose a global-to-local generation approach that mitigates the accumulated error and incorporates constraints among future steps.
\section{Methodology}

\subsection{Problem Formulation and Overview}\label{pf}
In the field of trajectory prediction, we are provided with the observed trajectory coordinates 
of an interested agent $a$, denoted as
$P=\left\{ \left(x_{i}, y_{i}\right)\right\}_{i=-T_p+1}^{{0}}$.
Here, $T_p$ represents the length of past steps.
Naturally, in the given scenario, agent $a$ interacts with various elements in the environment, including other agents and the map. For brevity, we denote these elements as $S$.
Our work can generally build upon any encoder-decoder trajectory predictor, denoted as $\mathcal{G}$.
Following the encoder-decoder structure of  $\mathcal{G}$, the agent features of $a$ can typically be derived by 
\begin{align}\label{fcx1}
  &{\mathcal{A}}={{\mathcal{G} }}\left( {{P}},{S} \right).
\end{align}
Through ${\mathcal{A}}$, the objective is to predict $a$'s future trajectory coordinates $F=\left\{ F_i\right\}_{i=1}^{{T_f}}$,  where $T_f$ is the length of future steps.
$F_{i}=\left(x_{i}, y_{i}\right)$ denotes the coordinate of step $t_i$.

The overview framework of our \sysname is depicted in \figref{framework}. 
Specifically, we propose a global-to-local generation approach for trajectory prediction.
In the example of \figref{framework}, 
initially, we simultaneously generate the global key steps $t_1$, $t_9$ and so on with a granularity of 8. 
Subsequently, the intermediate step $t_5$ is generated by utilizing $t_1$ and $t_9$.
Next, $t_3$ ($t_7$) is generated by utilizing $t_1$ and $t_5$ ($t_5$ and $t_9$).
This local recursion continues until all future steps are generated.

\subsection{Global Key Step Generation}\label{sp}
In this section, we generate key future time steps of agent $a$ at once.
Specifically, we set the granularity of key steps as $L$ (\ie the time interval between two adjacent key steps.)
Subsequently, we determine the key steps necessary to cover the length of future steps uniformly.
Let $\mathbb{G}_L$ denote the group of key steps under the granularity $L$, which can be derived by
\begin{align}\label{gl}
\mathbb{G}_L = \{t_1,t_{1+L},...,t_{1+NL}\},  \nonumber\\
s.t. \ \ \ \ 1+(N-1)L<T_f \leq 1+NL.
\end{align}
Here, we will generate a total of $N+1$ key steps.
Any two adjacent key steps are considered as the head and the tail steps of a local section.
Consequently, we partition the entire trajectory into $n$ sections.
In this manner, we ensure that the accumulated error among future steps would not propagate beyond each section.
Next, we predict the position coordinates $\hat{Z} \in \mathbb{R}^{(N+1)\times 2}$ of these $N+1$ key steps by
\begin{align}\label{m1}
\hat{Z} = \psi_L\left({\mathcal{A}}\right),
\end{align}
where $\psi_L$ is a two-layer Multilayer-Perceptron (MLP).
Let 
$\hat{Z}_{1+nL} \in \mathbb{R}^{2}$ denotes the corresponding predicted coordinate of step $t_{1+nL}$, where $n \in \{0,1,...,N\}$.

Since the key steps are generated simultaneously, these steps currently lack interrelated constraints, potentially causing kinematically infeasible predictions, as shown in \figref{overview}(a).
To address this issue,  we introduce spatial constraints among key steps by regulating their positional difference.
The detailed spatial constraint loss $\mathcal{L}_s^L$ under the granularity $L$ can be calculated by
\begin{equation}\label{lsp}
\resizebox{\linewidth}{!}{
    $\mathcal{L}_s^L=\sum\limits_{n=0}^{N-1}{{{\mathcal{L}}_{r}}}\left( \left( {\hat{Z}_{1+(n+1)L}}-{\hat{Z}_{1+nL}} \right),\left( {{F}_{1+(n+1)L}}-{{F}_{1+nL}} \right) \right)$,
}
\end{equation}
where $\mathcal{L}_{r}$ denotes the regression loss, such as Mean Square Error (MSE) loss \cite{gu2021densetnt} or the Negative Log-Likelihood (NLL) loss \cite{varadarajan2022multipath++}.
$F_{1+nL}$ denotes the ground truth coordinate of step $t_{1+nL}$.
Our \sysname adopts the MSE loss according to the empirical performance.
Notably, if $T_f < 1+NL$, the value $F_{1+NL}$ is not available.
In such cases, we set the upper limit of $n$ in \equref{lsp} to $N-2$.
As shown in \equref{lsp}, $\mathcal{L}_s^L$ endeavors to adjust inconsistent steps to a reasonable relative distance from adjacent steps.
In this way, we ensure generate consistent trajectories that adhere to kinematic feasibility principles.
\begin{figure}
\centering
\begin{overpic}[width=\linewidth]{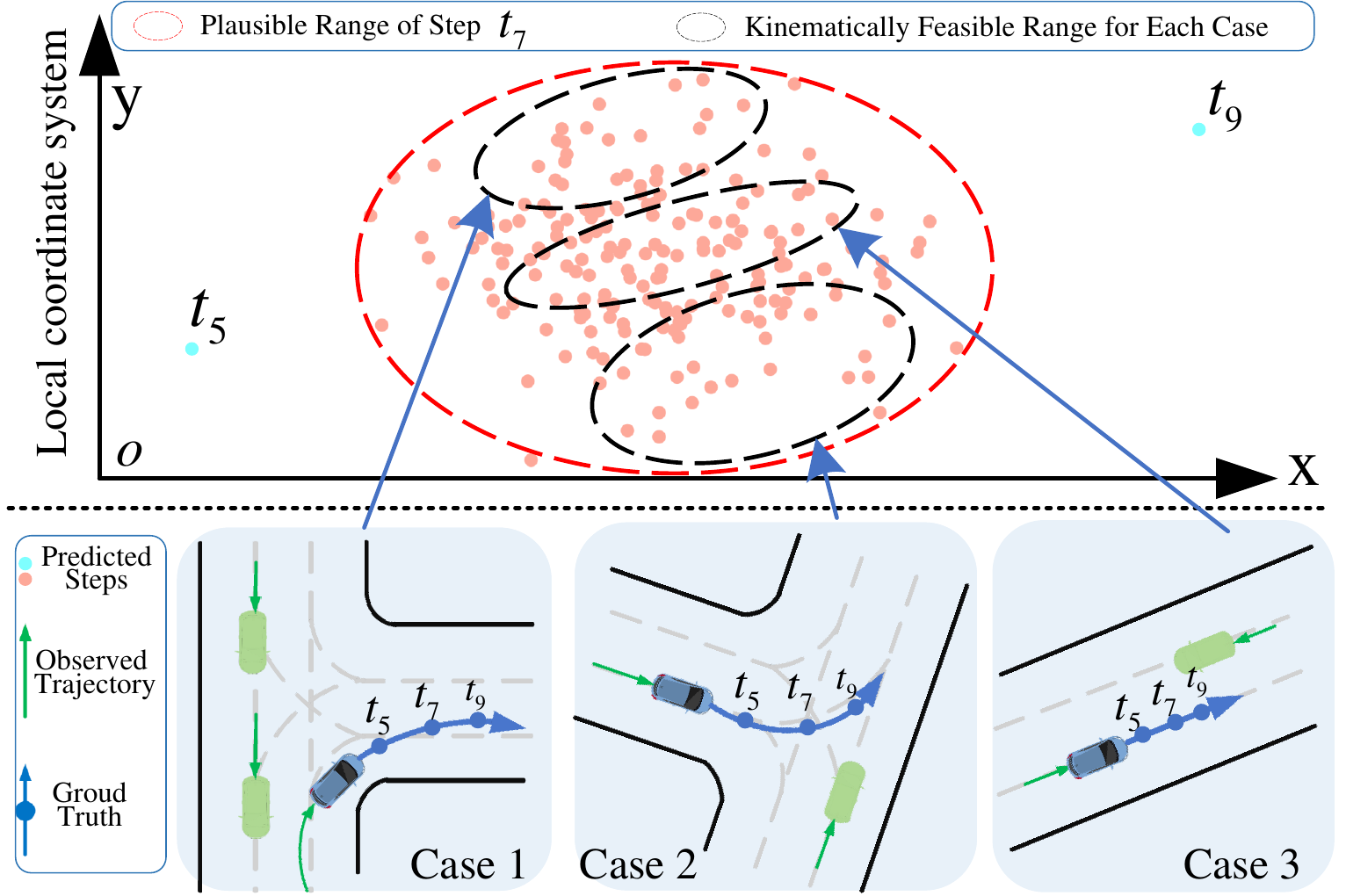}
\end{overpic}
\caption{Merely relying on $t_5$ and $t_9$ limits the model's access to historical kinematic information, possibly generating plausible but kinematically infeasible $t_7$, as shown in the red oval.
By incorporating historical kinematics clues, the prediction range can be reduced to the kinematically feasible range, as shown in the black oval.
}
\label{fig3}
\end{figure}
\begin{algorithm}[tb]
    \caption{Global-to-local Generation Process}
    \label{algorithm1}
    \textbf{Input}: The granularity of key steps $L$, the predicted key coordinates $\{{\hat{Z}_{1+iL}}\}_{i=0}^N$, two-layer MLPs $\{\phi_l\}_{l\in\{L,L/2,...,2\}}$, the position embedding $\{p_*\}_{*=0}^{1+NL}$, and agent features ${\mathcal{A}}$.\\
    \textbf{Output}: A predicted future trajectory $\hat{F}_L \in \mathbb{R}^{(1+NL)\times 2}$ for agent $a$.
    \begin{algorithmic}[1] 
        \STATE Let $l \leftarrow L$.
        \WHILE{$l>1$}
        \STATE Calculate $\left\{\hat{Z}_{1+\left(i+\frac{1}{2}\right)l}\right\}_{i=0}^{\frac{NL}{l}-1}$ by \equref{lg} in parallel with $\phi_l$, $\hat{Z}_{1+il}$, $p_{1+il}$,
        $\hat{Z}_{1+\left(i+1\right)l}$, $p_{1+\left(i+1\right)l}$ and ${\mathcal{A}}$ as input;
        \STATE $l \leftarrow l/2$.
        \STATE {$\triangleright$ $l$ \emph{denotes the current interval of existing adjacent predicted steps. 
        Notably, we set $L\in\{2,4,8,...\}$ to ensure $l/2$ is an integer value continuously.
        When $l=1$, it indicates that the entire trajectory coordinates of $1+NL$ steps have already been generated.}}
        \ENDWHILE
    \end{algorithmic}
\end{algorithm}
\subsection{Local Recursive Generation}\label{tem}
After generating the global key steps of agent $a$, the next phase is to fill in the intermediate steps within each local section.
If merely relying on the head and tail steps to generate the intermediate step, 
the temporal constraints among them are
fragile.
To illustrate this issue intuitively, we enumerate three cases, as shown in \figref{fig3}.
In these cases, the ground truths display similar $t_5$ and $t_9$ positions in the local coordinate system while exhibiting distinct positions for the intermediate step $t_7$.
Solely depending on $t_5$ and $t_9$ limits the model's access to historical kinematic information, resulting in delicate temporal constraints with past trajectories.
Consequently, this manner possibly generates plausible but kinematically infeasible $t_7$ which is inconsistent with past trajectories, as shown in the red oval in \figref{fig3}.

To alleviate this issue, we propose incorporating agent features in addition to the positional information of the head and tail steps.
Specifically, agent features ${\mathcal{A}}$ encompass historical kinematics clues.
By incorporating ${\mathcal{A}}$, we can improve the temporal constraints with past trajectories.
The improved constraints aim to reduce the plausible prediction range to the kinematically feasible range, as shown in the black oval in the \figref{fig3}.
Besides, we follow \cite{haoyietal-informer-2021,zeng2023transformers} to employ the position embedding of each step in the time sequence, as it can provide the local temporal information.
Next, given the predicted coordinates of the head step $t_i$ and the tail step $t_j$, the coordinate of the intermediate step $t_{(i+j)/2}$ can be derived by a projection layer
\begin{align}\label{lg1}
\hat{Z}_{(i+j)/2} =  \mathrm{Projection}\left(
\hat{Z}_i,p_i,{\hat{Z}_j},p_j,{\mathcal{A}}\right),
\end{align}
where $p_i\in \mathbb{R}^{D}$ and $p_j\in \mathbb{R}^{D}$ denote the position embeddings of steps $t_i$ and $t_j$, respectively.
Empirically, we project these elements in a simple yet effective manner, which can be calculated by
\begin{align}\label{lg}
\hat{Z}_{(i+j)/2} = \phi_l\left(\left({\hat{Z}_i}W_h+p_i\right)|| (\left({\hat{Z}_j}W_t+p_j\right)|| {\mathcal{A}}\right),
\end{align}
where $\phi_l$ is a two-layer MLP, $l$ is the interval between $t_i$ and $t_j$. 
Due to the changing interval, we utilize multiple $\phi_l$ functions that vary with the subscript $l$, instead of relying on a single MLP.
$||$ denotes the concatenation operation.
Since the values of ${\hat{Z}_i}$ and ${\hat{Z}_j}$
may differ significantly with a large $l$,
we project their dimension to align with that of the position embedding using two different weight matrices (\ie $W_h\in \mathbb{R}^{2\times {D}}$ and $W_t\in \mathbb{R}^{2\times {D}}$).
After finishing the global-to-local recursive generation process,
we obtain a predicted trajectory of agent $a$, denoted as $\hat{F}_L \in \mathbb{R}^{(1+NL)\times 2}$, as shown in Algorithm \ref{algorithm1}.
\begin{table*}[ht]
\setlength{\tabcolsep}{11.5pt}
\renewcommand{\arraystretch}{1.2}
\scalebox{0.85}{
\centering
\begin{tabular}{ccc|cc|cc|cc}
\specialrule{\heavyrulewidth}{0pt}{0pt}
    \multicolumn{1}{c}{\multirow{2}{*}{}} & \multicolumn{2}{c}{SGCN} & \multicolumn{2}{c|}{SGCN(G2LTraj)} & \multicolumn{2}{c}{STGCNN} & \multicolumn{2}{c}{STGCNN(G2LTraj)} \\
    \cline{2 - 9}
      & ADE$\downarrow$ & FDE$\downarrow$ & ADE$\downarrow$ (Gains)& FDE$\downarrow$ (Gains) & ADE$\downarrow$ & FDE$\downarrow$ & ADE$\downarrow$ (Gains)& FDE$\downarrow$ (Gains) \\ \specialrule{\heavyrulewidth}{0pt}{0pt}
     ETH & 0.360 & 0.565 & \textbf{0.340}(+5.56\%) & \textbf{0.548}(+3.01\%) & 0.365 & 0.582 & \textbf{0.362}(+0.82\%) & \textbf{0.576}(+1.03\%) \\
     HOTEL & 0.131 & 0.210 & \textbf{0.123}(+6.11\%) & \textbf{0.197}(+6.19\%) & 0.147 & 0.220 & \textbf{0.133}(+9.52\%) & \textbf{0.211}(+4.09\%) \\
     UNIV & 0.244 & \textbf{0.430} & \textbf{0.242}(+0.82\%) & 0.431(-0.70\%) & 0.246 & \textbf{0.427} & \textbf{0.244}(+0.81\%) & 0.430(-0.70\%)\\ 
     ZARA1 & 0.200 & \textbf{0.347} & \textbf{0.193}(+3.50\%) & 0.349(-0.58\%) & 0.217 & 0.393 & \textbf{0.214}(+1.38\%) & \textbf{0.380}(+3.31\%) \\
     ZARA2 & 0.153 & 0.261 & \textbf{0.147}(+3.92\%) & \textbf{0.255}(+2.30\%) & 0.168 & 0.290 & \textbf{0.160}(+4.76\%) & \textbf{0.277}(+4.48\%) \\ \specialrule{\heavyrulewidth}{0pt}{0pt}
     \noalign{\vskip 4pt} \specialrule{\heavyrulewidth}{0pt}{0pt}
     \multicolumn{1}{c}{\multirow{2}{*}{}} & \multicolumn{2}{c}{PECNet} & \multicolumn{2}{c|}{PECNet(G2LTraj)} & \multicolumn{2}{c}{AgentFormer} & \multicolumn{2}{c}{AgentFormer(G2LTraj)} \\
     \cline{2 - 9}
      & ADE$\downarrow$ & FDE$\downarrow$ & ADE$\downarrow$ (Gains)& FDE$\downarrow$ (Gains)& ADE$\downarrow$ & FDE$\downarrow$ & ADE$\downarrow$ (Gains)& FDE$\downarrow$ (Gains)\\ \specialrule{\heavyrulewidth}{0pt}{0pt}
      ETH & 0.365 & 0.572 & \textbf{0.348}(+4.66\%) & \textbf{0.536}(+6.29\%) & 0.362 & \textbf{0.568} & \textbf{0.354}(+2.21\%) & 0.592(-4.23\%) \\
      HOTEL & 0.132 & 0.211 & \textbf{0.116}(+12.12\%) & \textbf{0.183}(+13.27\%) & 0.147 & \textbf{0.222} & \textbf{0.136}(+7.48\%) & 0.228(-2.70\%) \\
      UNIV & 0.244 & 0.432 & \textbf{0.241}(+1.23\%) & \textbf{0.424}(+1.85\%) & 0.244 & 0.430 & \textbf{0.234}(+4.10\%) & \textbf{0.416}(+3.26\%) \\
      ZARA1 & 0.195 & 0.348 & \textbf{0.185}(+5.13\%) & \textbf{0.329}(+5.46\%) & 0.216 & 0.397 & \textbf{0.211}(+2.31\%) & \textbf{0.389}(+2.02\%) \\
      ZARA2 & 0.143 & 0.250 & \textbf{0.138}(+3.50\%) & \textbf{0.241}(+3.60\%) & 0.166 & 0.290 & \textbf{0.163}(+1.81\%) & \textbf{0.282}(+2.76\%) \\ \specialrule{\heavyrulewidth}{0pt}{0pt}
     \noalign{\vskip 4pt} \specialrule{\heavyrulewidth}{0pt}{0pt}
     \multicolumn{1}{c}{\multirow{2}{*}{}} & \multicolumn{2}{c}{GraphTERN} & \multicolumn{2}{c|}{GraphTERN(G2LTraj)} & \multicolumn{2}{c}{DMRGCN} & \multicolumn{2}{c}{DMRGCN(G2LTraj)} \\
     \cline{2 - 9}
      & ADE$\downarrow$ & FDE$\downarrow$ & ADE$\downarrow$ (Gains) & FDE$\downarrow$ (Gains)& ADE$\downarrow$ & FDE$\downarrow$ & ADE$\downarrow$ (Gains)& FDE$\downarrow$ (Gains)\\ \specialrule{\heavyrulewidth}{0pt}{0pt}
      ETH & 0.378 & \textbf{0.578} & \textbf{0.360}(+4.76\%) & 0.589(-1.90\%) & 0.382 & 0.603 & \textbf{0.366}(+4.19\%) & \textbf{0.577}(+4.31\%) \\
      HOTEL & 0.129 & 0.205 & \textbf{0.122}(+5.43\%) & \textbf{0.196}(+4.39\%) & 0.129 & 0.204 & \textbf{0.124}(+3.88\%) & \textbf{0.197}(+3.43\%) \\
      UNIV & 0.255 & 0.447 & \textbf{0.248}(+2.75\%) & \textbf{0.434}(+2.91\%) & 0.256 & 0.439 & \textbf{0.245}(+4.30\%) & \textbf{0.427}(+2.73\%) \\
      ZARA1 & 0.209 & 0.373 & \textbf{0.204}(+2.39\%) & \textbf{0.366}(+1.88\%) & 0.210 & 0.377 & \textbf{0.207}(+1.43\%) & \textbf{0.369}(+2.12\%) \\
      ZARA2 & 0.157 & 0.273 & \textbf{0.156}(+0.64\%) & \textbf{0.262}(+4.03\%) & 0.165 & 0.284 & \textbf{0.159}(+3.64\%) & \textbf{0.275}(+3.17\%) \\ \specialrule{\heavyrulewidth}{0pt}{0pt}
\end{tabular}
}
\caption{Performance comparison between six trajectory prediction predictors and our \sysname on ETH and UCY test set.
We obtain the performance values of SGCN, STGCNN, PECNet and AgentFormer from the original paper \protect\cite{bae2023eigentrajectory}, while the performance values of GraphTERN, and DMRGCN are derived based on their official codes.
For a fair comparison with these state-of-the-art methods, all predictors are constructed based on the Eigentrajectory space \protect\cite{bae2023eigentrajectory}.
Best performances are in \textbf{bold}.
Gains: performance improvements, Unit for ADE and FDE: meter.
}
\label{tab1}
\end{table*}
\subsection{Selectable Granularity Strategy}\label{stra}
Future Trajectories of different agents can exhibit diverse kinematics characteristics.
Thus, employing a fixed granularity may not be appropriate for each trajectory.
For instance, unusual behaviors like abrupt direction changes involve rapid shifts in velocity and trajectory. In such cases, a finer granularity might be more adequate as it can better capture sudden motion changes. 
Conversely, for usual behaviors such as moving straight uniformly, a coarser granularity might suffice, as it can enhance the temporal constraints from more distant steps. 
To this end, we propose a selectable granularity strategy that aims to be as suitable as possible for each trajectory.
In the following, for the sake of brevity, we focus on discussing the cases where $T_f = 2^M +1$, $M$ is an integer.

The selectable strategy begins by generating key steps at a fine granularity.
In other words, we set $L$ to 2 and obtain the key group $\mathbb{G}_{L=2}=\{t_1,t_{3},...,t_{T_f}\}$,
Subsequently, we downsample $\mathbb{G}_{L=2}$ to derive $M-1$ groups of coarse-grained key steps, \ie $\mathbb{G}_{L=4}=\{t_1,t_{5},...,t_{T_f}\}$, ..., $\mathbb{G}_{L={2^M}}=\{t_1,t_{T_f}\}$.
Based on these groups, we follow the above procedure to generate $M$ trajectories, denoted as $\{\hat{F}_2,\hat{F}_4,...,\hat{F}_{2^M}\}$.
To select the optimal trajectory from these $M$ trajectories,
 we learn their confidence scores $\hat{C} \in \mathbb{R}^{M}$ by 
\begin{align}\label{cs}
\hat{C} = \sigma\left(\psi_c\left(\hat{Z}_1 || \hat{Z}_{T_f} || {\mathcal{A}}\right)\right),
\end{align}
where $\psi_c$ is a two-layer MLP, $\sigma$ denotes a softmax function. 
$\hat{Z}_1$ and $\hat{Z}_{T_f}$ are the shared head and tail coordinates of these $M$ trajectories.
By incorporating them with $\mathcal{A}$, the model can capture the overall trajectory outlines, thereby enhancing the evaluation of the $M$ trajectories.
Subsequently, we establish the ground truth for the confidence scores of $\{\hat{F}_{2^m}\}_{m=1}^M$ by
\begin{align}\label{gt}
{C_{m}}={\frac{\mathrm{exp}\left(-\mathrm{ADE}(\hat{F}_{2^m},{F})\right)}{\sum\limits_{j=1}^M{\mathrm{exp}\left(-\mathrm{ADE}(\hat{F}_{2^j},{F})\right)}}}, 
\end{align}
where $\mathrm{ADE}$ denotes the
average L2 distance between each coordinate of the predicted
trajectory and the ground truth. 
Let $C$ denote $\{C_m\}_{m=1}^M$. 
The predicted confidence scores $\hat{C}$ can be optimized by
\begin{align}\label{lc}
\mathcal{L}_c=\mathcal{L}_{r}\left( {\hat{C}}, C \right),
\end{align}
where $\mathcal{L}_{r}$ denotes the regression loss.
Notably, during the inference process, 
we only generate the predicted trajectory corresponding to the highest confidence score through our global-to-local approach.
This way can significantly reduce the inference time.
The final loss function can be denoted as
\begin{align}\label{final}
\mathcal{L}=\mathcal{L}_{\mathcal{G}} + \eta_1 \sum\limits_{i=1}^M
\mathcal{L}_s^{2^i} + \eta_2 \mathcal{L}_c,
\end{align}
where $\mathcal{L}_{\mathcal{G}}$ denotes the inherent loss of the existing predictor ${\mathcal{G}}$,
$\mathcal{L}_s^{2^i}$ is the spatial constraint loss in \equref{lsp},
$\eta_1$ and $\eta_2$  are trade-off hyper-parameters.
\section{Experiments}
\subsection{Experimental Setup}
{\setlength{\parindent}{0cm}\textbf{Datasets.}}
We conducted experiments on three widely-used datasets: ETH \cite{pellegrini2009you}, UCY \cite{lerner2007crowds} and nuScenes \cite{caesar2020nuscenes}.
ETH and UCY comprise five distinct real-world scenarios, (\ie ETH and HOTEL from ETH; UNIV, ZARA1 and ZARA2 from UCY), where 1,536 pedestrians were recorded in the world coordinates. 
We employ the standard leave-one-out strategy \cite{bae2023eigentrajectory,alahi2016social} for training and evaluation.
The task in ETH and UCY is to predict the future 12 steps (4.8 seconds) based on the observed 8 steps (3.2 seconds).
nuScenes is a large-scale trajectory dataset that includes both vehicles and pedestrians.
The task in nuScenes is to predict the future 12 steps (6 seconds) given the observed 4 steps (2 seconds) and the High-Definition (HD) maps.

{\setlength{\parindent}{0cm}\textbf{Baseline Predictors.}}
We evaluate the effectiveness of our \sysname by incorporating it into seven state-of-the-art predictors: SGCN \cite{shi2021sgcn}, STGCNN \cite{mohamed2020social}, PECNet \cite{mangalam2020not}, AgentFormer \cite{yuan2021agentformer}, GraphTERN \cite{bae2023set}, and DMRGCN \cite{bae2021disentangled} for the ETH and UCY datasets, and AutoBot \cite{girgis2022latent} for the nuScenes datasets. 

{\setlength{\parindent}{0cm}\textbf{Evaluation Metrics.}}
Following the standard evaluation strategy \cite{bae2023eigentrajectory,zhao2021you}, we assess the performance on ETH and UCY using the average displacement error(ADE) and the final displacement error(FDE). 
ADE calculates the average L2 distance between all predicted trajectory coordinates and the corresponding ground truth coordinates. 
FDE  measures the L2 distance between the final predicted destination coordinates and those of the ground truth.
minADE$_5$ and minADE$_{10}$ represent the average pointwise L2 distances between the predicted trajectory and the ground truth for the 5 and 10 most likely predictions, respectively,
minFDE$_1$ denotes the L2 distance between the final points of the prediction and the ground truth of the most likely
prediction.
MR$_5$ ( MR$_{10}$ ) calculates the ratio of
scenarios where none of the predicted 5 (10) trajectories are within 2 meters of the ground truth according to FDE.
We use the best-of-$K$ strategy to compute the final ADE and FDE with $K$=20, following prior works.
We evaluate the results on nuScenes validation set with $K$=10, following \cite{girgis2022latent}.

{\setlength{\parindent}{0cm}\textbf{Implementation Details.}}
 We set the position embedding dimension $D$  as 64 in Sec. \ref{tem}.
For each trajectory, We select the optimal granularity from three groups, \ie  $\mathbb{G}_{L=2}$, $\mathbb{G}_{L=4}$ and $\mathbb{G}_{L=8}$.
In \equref{final}, $\eta_1$ and $\eta_2$ are set as 0.1 and 1, respectively.
For a fair comparison, we adopt the default configurations of the baseline predictors \cite{bae2023eigentrajectory,girgis2022latent}.
Specifically, for ETH and UCY, we employ the AdamW optimizer \cite{loshchilov2018decoupled}, set the batch size to 128, use a learning rate of 0.001, and train for 256 epochs. For nuScenes, we utilize the Adam optimizer \cite{kingma2014adam}, set the batch size to 64, use a learning rate of 7.5e$^{-4}$, and train for 80 epochs. The model is trained using a single 3090 GPU.
Notably, in practice,  the predictor $\mathcal{G}$ commonly generates $K$ trajectories for each agent, aiming to model the multimodal distribution of future trajectories \cite{bae2023set,girgis2022latent,bae2023eigentrajectory}.
The global-to-local generation process of these trajectories is the same and in parallel.

\subsection{Performance Comparison of Main Results}\label{mainre}
{\setlength{\parindent}{0cm}\textbf{Performance Comparison on ETH and UCY.}}
As shown in \tabref{tab1}, 
our \sysname framework consistently outperforms all the baseline models in most cases. 
Notably, when compared to the simultaneous methods, our approach achieves substantial improvements in ADE and FDE, reaching up to 12.12\% and 13.27\%. 
Similarly, when compared to the recursive method AgentFormer, our approach demonstrates a significant enhancement in prediction accuracy. 
Our \sysname also has certain limitations, particularly when evaluating on the UNIV subset, using SGCN and STGCNN as the backbone predictors.
Overall, these results highlight the effectiveness of our global-to-local generation approach.
\begin{table}[!t]
\setlength{\tabcolsep}{0pt}
\renewcommand{\arraystretch}{1.2}
\centering
\scalebox{0.85}{
    \begin{tabular}{c|ccccc}
        \specialrule{\heavyrulewidth}{0pt}{0pt}
        Model  & minADE$_5$$\downarrow$ & minADE$_{10}$$\downarrow$  & MR$_5$$\downarrow$
        & MR$_{10}$$\downarrow$ & minFDE$_1$$\downarrow$ 
        \\
        \specialrule{\heavyrulewidth}{0pt}{0pt}
        AutoBot$_{NM}$ & 1.88 & 1.35 & 0.72 & 0.58
        & - \\
        AutoBot$_{NM}$(\sysname)  & \textbf{1.77} & \textbf{1.24} & {0.72} & \textbf{0.55}
        & -  \\
        \specialrule{\heavyrulewidth}{0pt}{0pt}
        AutobBot  & 1.43 & 1.05 &0.66 & 0.45
        & 8.66 \\
        AutoBot(\sysname)  & \textbf{1.40} & \textbf{0.96} & \textbf{0.63} & \textbf{0.39}
        & \textbf{8.30} \\
        \specialrule{\heavyrulewidth}{0pt}{0pt}
        AutoBot$_{EN}$  & 1.37 & 1.03 & 0.62 & 0.44
        & 8.19 \\
        AutoBot$_{EN}$(\sysname)  & \textbf{1.32} & \textbf{0.90} & 0.62 & \textbf{0.37}
        & \textbf{8.01} \\
        \specialrule{\heavyrulewidth}{0pt}{0pt}
    \end{tabular}
}
\caption{Performance comparison between AutoBot and our \sysname on three benchmarks of nuScenes.
Compared with AutoBot, AutoBot$_{NM}$ denotes that the model trained without incorporating the map information.
AutoBot$_{EN}$ denotes the ensemble of three AutoBot models to enhance final predictions. 
We obtain the performance values of AutoBot$_{NM}$, AutoBot and AutoBot$_{EN}$ from the original paper \protect\cite{girgis2022latent}.
Best performances are in \textbf{bold}.}
\label{tab2}
\end{table}

{\setlength{\parindent}{0cm}\textbf{Performance Comparison on nuScenes.}}
As shown in \tabref{tab2}, 
our \sysname framework continuously surpasses AutoBot in three benchmarks in most cases. 
For example, when utilizing the ensemble strategy, \sysname achieves a significant decrease of 3.65\% and 12.6\% in terms of minADE$_5$ and minADE$_{10}$.
On the other hand, as for the MR$_5$ metric, the performance of \sysname exhibits a slight improvement compared to the raw predictor.
Notably, the ensemble strategy demonstrates the best performance across all metrics.
whereas the model without map information exhibits the poorest performance.

\subsection{Ablation Studies}
\begin{table}[!t]
\setlength{\tabcolsep}{4pt}
\renewcommand{\arraystretch}{1.2}
\centering
\scalebox{0.85}{
    \begin{tabular}{c|ccccc}
        \specialrule{\heavyrulewidth}{0pt}{0pt}
        Model  & minADE$_5$$\downarrow$ & minADE$_{10}$$\downarrow$  & MR$_5$$\downarrow$
        & MR$_{10}$$\downarrow$ & minFDE$_1$$\downarrow$ 
        \\
        \specialrule{\heavyrulewidth}{0pt}{0pt}
        \sysname-R  & 1.61 & 1.03 & 0.68 & 0.48
        & 9.41 \\
        \sysname  & \textbf{1.40} & \textbf{0.96} & \textbf{0.63} & \textbf{0.39}
        & \textbf{8.30} \\
        \specialrule{\heavyrulewidth}{0pt}{0pt}
    \end{tabular}
}
\caption{Effectiveness analysis in mitigating the accumulated error.
Instead of generating key steps, the \sysname-R variant simultaneously generates all future steps and uses an RNN-based network to refine them, following \protect\cite{jia2023towards}.
}
\label{tab3}
\end{table}
We conduct several ablation studies
based on the AutoBot predictor and evaluated on nuScenes,
unless otherwise stated.

\begin{figure}
\centering
\begin{overpic}[width=\linewidth]{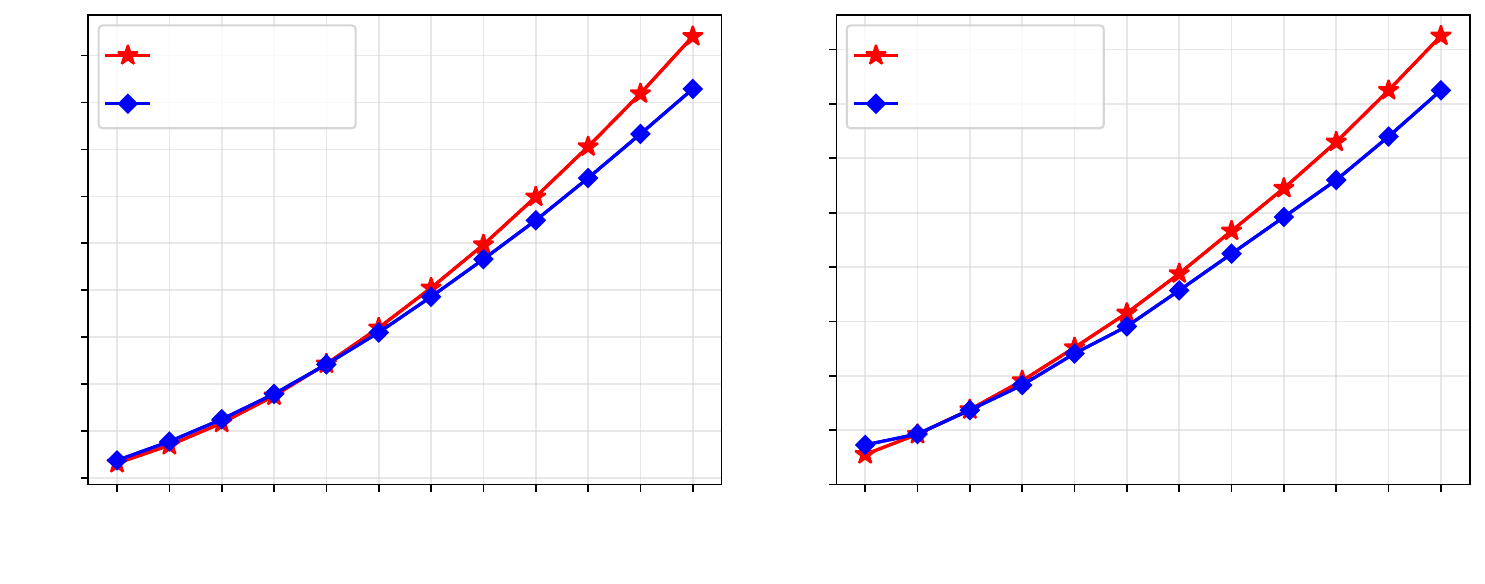}
\end{overpic}
\caption{The accumulated error comparison at each future step.
Accumulated Error (5) and (10)
are the average of pointwise L2 distance between each future step and ground truth over
5 and 10 most likely predictions respectively. }
\label{fig4}
\end{figure}

{\setlength{\parindent}{0cm}\textbf{Effectiveness for Mitigating the Accumulated Error.}}
In this part, we validate the effectiveness for mitigating the accumulated error.
Specifically, as shown in \tabref{tab3}, 
compared to \sysname, \sysname-R exhibits inferior performance. 
We show the comparison results more intuitively in \figref{fig4}.
Notably, \sysname-R outperforms our \sysname by a margin when the time step is less than 3.
This presents the effectiveness of the RNN-based network in successfully incorporating temporal constraints for short-term trajectory prediction.
As the time step increases, both approaches experience a growth in the accumulated error.
However, \sysname exhibits a slower growth trend compared to \sysname-R. 
It demonstrates that \sysname effectively 
mitigates the accumulated error by partitioning the entire trajectory into multiple sections.
\begin{table}[!t]
\setlength{\tabcolsep}{4pt}
\renewcommand{\arraystretch}{1.2}
\centering
\scalebox{0.85}{
    \begin{tabular}{c|ccccc}
        \specialrule{\heavyrulewidth}{0pt}{0pt}
        Model  & minADE$_5$$\downarrow$ & minADE$_{10}$$\downarrow$  & MR$_5$$\downarrow$
        & MR$_{10}$$\downarrow$ & minFDE$_1$$\downarrow$ 
        \\
        \specialrule{\heavyrulewidth}{0pt}{0pt}
        w/o SC  & 1.45 & 1.03 & 0.63 & 0.44
        & 8.51 \\
        SC- CG  & 1.42 & 0.98 & 0.65 & 0.43
        & {8.49} \\
        SC-FG  & 1.42 & 0.96 & 0.65 & 0.40
        & {8.73} \\
        \sysname  & \textbf{1.40} & \textbf{0.96} & \textbf{0.63} & \textbf{0.39}
        & \textbf{8.30} \\
        \specialrule{\heavyrulewidth}{0pt}{0pt}
    \end{tabular}
}
\caption{Effectiveness analysis of spatial constraints.
w/o SC denotes removing all spatial constraints.
SC- CG and SC- FG denote imposing the spatial constraints only under coarse granularities (\ie $L>2$) and the fine granularity (\ie $L=2$), respectively.
}
\label{tab4}
\end{table}

{\setlength{\parindent}{0cm}\textbf{Effectiveness Analysis of Spatial Constraints.}}
To evaluate the efficacy of spatial constraints in \secref{sp}, we devise three variants. 
As shown in \tabref{tab4}, 
without any spatial constraint, the model exhibits the lowest performance, resulting in reductions of 3.6\% and 7.3\% in terms of minADE$_5$ and minADE${_{10}}$, respectively.
Besides, 
the constraints applied at the fine granularity level outweigh those applied at the coarse granularity level.
Both constraints at the fine and coarse granularity level contribute to performance gains.

{\setlength{\parindent}{0cm}\textbf{Effectiveness Analysis of Temporal Constraints.}}
To assess the effectiveness of temporal constraints in \secref{tem}, we introduce four different variants. 
As shown in \figref{tab5}, 
without the position embedding, the model 
performance slightly decreases.
Without agent features providing the past kinematics information, the model suffers from a substantial performance reduction of approximately 3.13\% in terms of minADE$_{10}$.
Only relying on the head and the tail steps, the model is limited to the fragile temporal constraints, resulting in the lowest performance.

\begin{table}[!t]
\setlength{\tabcolsep}{4pt}
\renewcommand{\arraystretch}{1.2}
\centering
\scalebox{0.85}{
    \begin{tabular}{c|ccccc}
        \specialrule{\heavyrulewidth}{0pt}{0pt}
        Model  & minADE$_5$$\downarrow$ & minADE$_{10}$$\downarrow$  & MR$_5$$\downarrow$
        & MR$_{10}$$\downarrow$ & minFDE$_1$$\downarrow$
        \\
        \specialrule{\heavyrulewidth}{0pt}{0pt}
        w/o PE  & 1.41 & 0.98 & 0.63 & 0.41
        & 8.35  \\
        w/o $\mathcal{A}$  & 1.43 & 0.99 & 0.65 & 0.43
        & {8.53} \\
        only-HT  & 1.43 & 1.02 & 0.64 & 0.42
        & {8.57} \\
        \sysname-I  & 1.43 & 1.01 & 0.65 & 0.45
        & {8.73} \\
        \sysname  & \textbf{1.40} & \textbf{0.96} & \textbf{0.63} & \textbf{0.39}
        & \textbf{8.30}\\
        \specialrule{\heavyrulewidth}{0pt}{0pt}
    \end{tabular}
}
\caption{Ablation studies of temporal constraints.
w/o PE and w/o A denote removing position embeddings and agent features in {\protect\equref{lg}}, respectively.
only-HT denotes relying on the head and tail steps to generate the intermediate step.
\sysname-I denotes generating the intermediate steps within each section at once.}
\label{tab5}
\end{table}

\begin{table}[!t]
\setlength{\tabcolsep}{4pt}
\renewcommand{\arraystretch}{1.2}
\centering
\scalebox{0.85}{
    \begin{tabular}{c|ccccc}
        \specialrule{\heavyrulewidth}{0pt}{0pt}
        Model  & minADE$_5$$\downarrow$ & minADE$_{10}$$\downarrow$  & MR$_5$$\downarrow$
        & MR$_{10}$$\downarrow$ & minFDE$_1$$\downarrow$
        \\
        \specialrule{\heavyrulewidth}{0pt}{0pt}
        $\mathbb{G}_{L=2}$  & 1.42 & 1.03 & 0.65 & 0.41
        & \textbf{8.27} \\
        $\mathbb{G}_{L=4}$  & 1.43 & 1.01 & 0.64 & 0.40
        & {8.48}\\
        $\mathbb{G}_{L=8}$ & 1.43 & 1.03 & 0.65 & 0.42
        & {8.60} \\
        \sysname  & \textbf{1.40} & \textbf{0.96} & \textbf{0.63} & \textbf{0.39}
        & \textbf{8.30}\\
        \specialrule{\heavyrulewidth}{0pt}{0pt}
    \end{tabular}
}
\caption{Ablation studies of the selectable granularity strategy.
$\mathbb{G}_{L=2}$, $\mathbb{G}_{L=4}$, $\mathbb{G}_{L=8}$ denote 
using a fixed granularity (\ie 2,4 and 8, respectively) when generating all future trajectories.
}
\label{tab6}
\end{table}

\begin{figure}
\centering
\begin{overpic}[width=\linewidth]{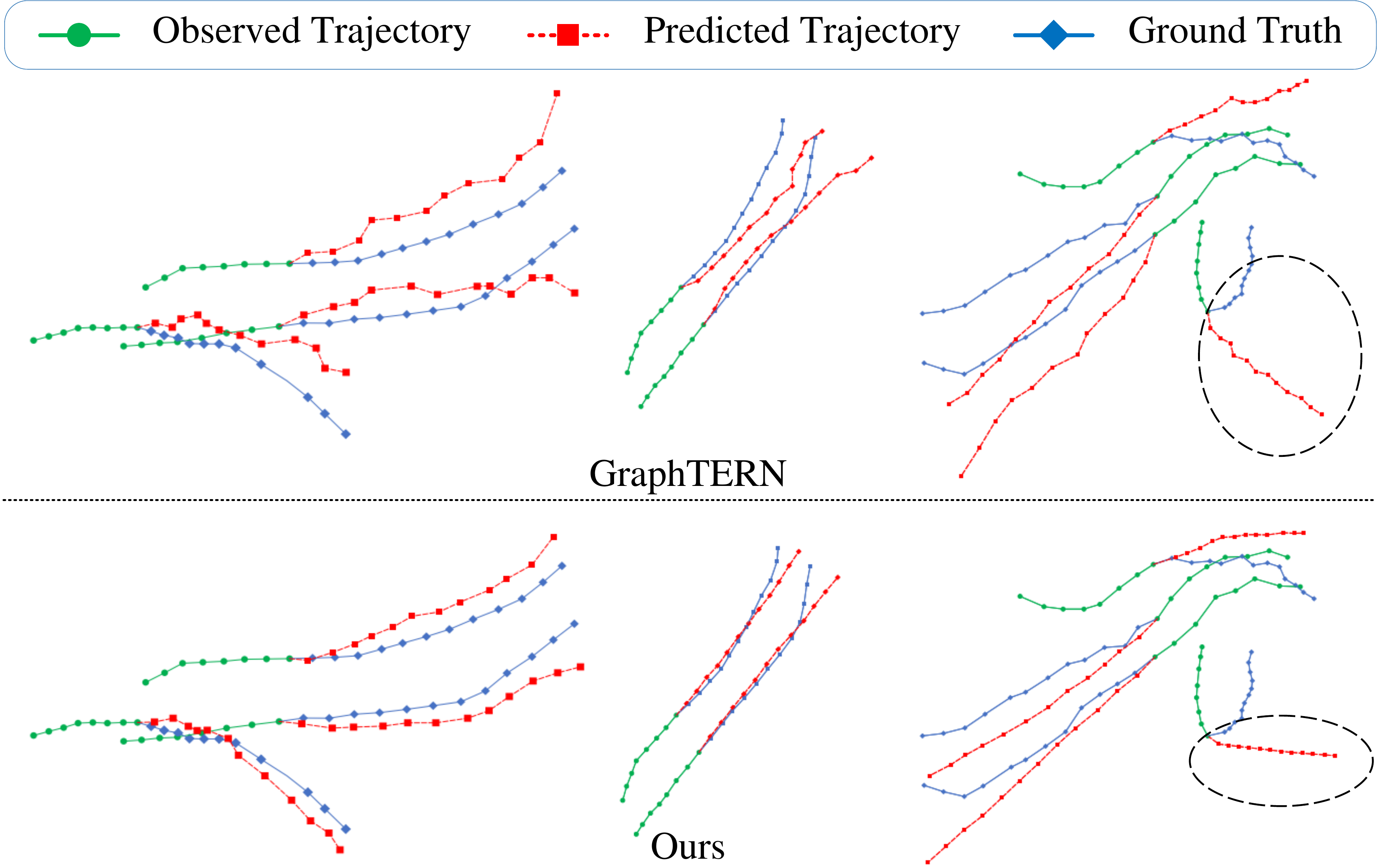}
\end{overpic}
\caption{Qualitative results on the UCY validation set. 
The black ovals denote a failure case.
}
\label{fig5}
\end{figure}

{\setlength{\parindent}{0cm}\textbf{Effectiveness Analysis of the Selectable Granularity Strategy.}}
In this part, we examine the effectiveness of our selectable granularity strategy in \secref{stra}.
As shown in \tabref{tab6}, employing three distinct fixed granularities consistently results in a reduction in performance, while the performance of $\mathcal{G}_2$ slightly surpasses that of \sysname.
Overall, the prediction performance across these three granularities surpasses that of the raw predictor. 
This finding confirms the efficacy of the global-to-local generation approach.

{\setlength{\parindent}{0cm}\textbf{Qualitative Results.}}
To illustrate the performance more intuitively, we randomly select several scenarios from the UCY \cite{lerner2007crowds} validation set.
As depicted in \figref{fig5}, the results clearly illustrate that \sysname greatly enhances the prediction performance compared to GraphTERN \cite{bae2023set}. Specifically, when spatial-temporal constraints are not enforced among future steps, the predicted trajectories from GraphTERN exhibit more pronounced fluctuations compared to \sysname. 
In contrast, the predicted trajectories generated by \sysname demonstrate greater consistency. 
This confirms that our \sysname tends to produce kinematically feasible predictions. 
Additionally, the predicted trajectories from our \sysname align more closely with the ground truth. 
Notably, both methods face challenges in accurately modeling unexpected behaviors, such as abrupt changes in direction, as shown in the black oval in \figref{fig5}. 
These failure cases generally involve rapid shifts in velocity and trajectory.

\section{Conclusion}
In this paper, we introduce a global-to-local generation approach for trajectory prediction that alleviates the accumulated error and introduces the constraints among future steps.
To enhance the kinematical feasibility, we propose the spatial constraints among the global key steps and boost the temporal constraints among the local intermediate steps.
To ensure the optimal granularity of key steps for each trajectory, we introduce a selectable granularity strategy. 
Our \sysname achieves noteworthy performance improvements over seven existing trajectory predictors when evaluated on three widely used datasets.
Experiment results verify its effectiveness.

{\setlength{\parindent}{0cm}\textbf{Acknowledgement.}}
This work was supported in part by The National Nature Science Foundation of China (Grant No: 62303406, 62273302, 62273303, 62036009, 61936006),  in part by Yongjiang Talent Introduction Programme (Grant No: 2023A-194-G, 2022A-240-G).

\bibliographystyle{named}
\bibliography{ijcai24}

\newpage
\setcounter{section}{0}
\renewcommand\thesection{\Alph{section}}

Considering the space limitation of the main text, we provided more results and discussion in this supplementary material, which is organized as follows: 
\begin{itemize}
\item Section \ref{aid}: Additional Implementation Details.
\item Section \ref{aad}: Additional Analysis and Discussions.
\begin{itemize}
\item Section \ref{sec1}:  Latency Analysis.
\item Section \ref{hy}:  Sensitivity Analysis of Hyper-parameters.
\item Section \ref{lgp}:  Analysis of Local Recursive Generation.
\item Section \ref{rl}:  Analysis of The Regression Loss.
\item Section \ref{kf}: Kalman Filtering for Post-processing.
\item Section \ref{sgs}: Visualization of The Selectable Granularity Strategy.

\end{itemize}
\end{itemize}

\section{Additional Implementation Details}\label{aid}
\begin{figure}
\centering
\renewcommand{\thefigure}{S1}
\begin{overpic}[width=\linewidth]{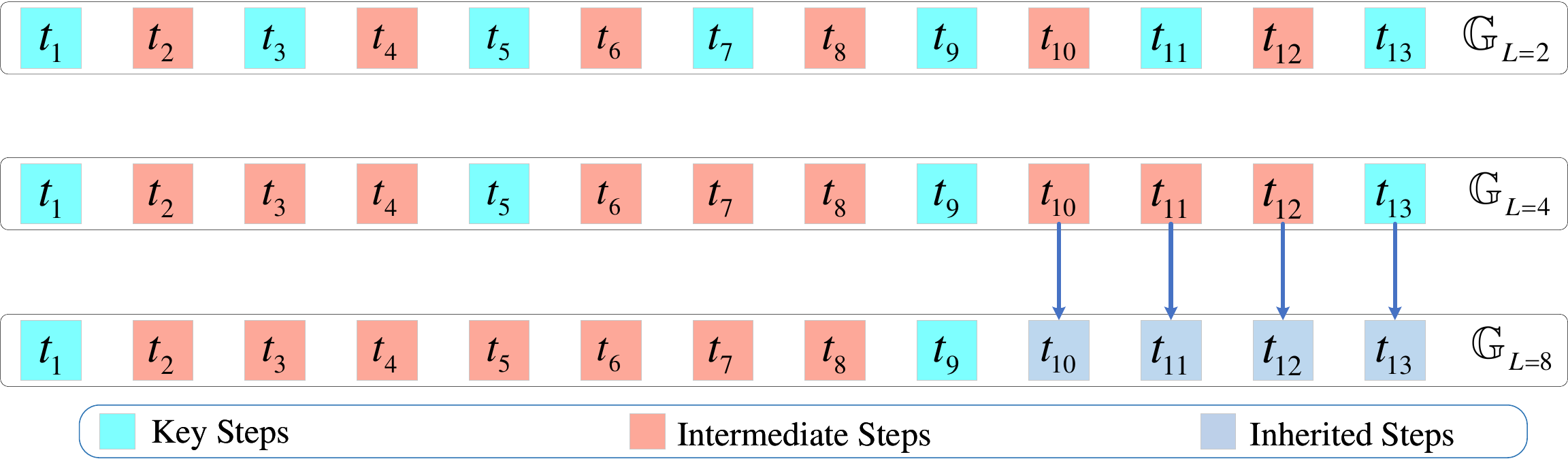}
\end{overpic}
\caption{During the global-to-local generation process, 
$t_{10}$, $t_{11}$, $t_{12}$, and $t_{13}$ of the group $\mathbb{G}_{L={8}}$ are unavailable.
These steps are inherited from the group  $\mathbb{G}_{L={4}}$.}
\label{tf}
\end{figure}

In practice, the length of future steps $T_f$ might not be equal to $2^M+1$, where $M$ is an integer.
For instance, in our experiments,
$T_f$ is set to 12.
In this instance,
as shown in \figref{tf}, we set the granularity $L$ to 2 and obtain the key group $\mathbb{G}_{L=2}=\{t_1,t_{3},...,t_{13}\}$.
Next, we downsample $\mathbb{G}_{L=2}$ to derive coarse-grained key steps, \ie $\mathbb{G}_{L=4}=\{t_1,t_{5},t_9,t_{13}\}$, ..., $\mathbb{G}_{L={8}}=\{t_1,t_{9}\}$.
As for $\mathbb{G}_{L={8}}$, the last key step along the timeline is $t_9$, resulting in $t_{10}$, $t_{11}$, $t_{12}$, and $t_{13}$ being unavailable.
In such cases, these steps are inherited from the group with adjacent smaller granularity (\ie $\mathbb{G}_{L={4}}$).
Notably, the length of the predicted trajectories is 13.
However, the ground truth coordinate $F_{13}=\left(x_{13}, y_{13}\right)$ of step $t_{13}$ is not available.
To maintain length consistency while optimizing the loss in  Eq. (10) in our main paper,
we approximate the ground truth value of $F_{13}$ by
\begin{equation}\label{t13}\tag{S1}
F_{13} = \left(F_{12} - F_{11}\right) +  F_{12}
\end{equation}

\section{Additional Analysis and Discussions}\label{aad}
We perform additional studies and discussions to gain further insight into \sysname. Unless otherwise stated, these studies are based on the AutoBot predictor \cite{girgis2022latent} and evaluated on the nuScenes dataset \cite{caesar2020nuscenes}.
\subsection{Latency Analysis}\label{sec1}
Since our \sysname builds upon AutoBot, we introduce a global-to-local generation head while preserving its original encoder-decoder structure.
Let $O\left(\mathcal{G}\right)$ denote the inference computation complexity of AutoBot. 
During inference, we only generate the predicted trajectory with the highest confidence score.
As a result, the computation complexity of our \sysname at the under the granularity $L$ can be approximated as $O\left(\mathcal{G}\right)$+$\sum\limits_{j=1}^{\mathrm{log}L}O\left(l_j\right)$+$O\left(c\right)$,
where $O\left(l_j\right)$ denotes the computation complexity of the $j$-th local recursive process, and
$O\left(c\right)$ denotes the computation complexity of the granularity selection process.
In practice, $\sum\limits_{j=1}^{\mathrm{log}L}O\left(l_j\right)+O\left(c\right) $ is significantly lower than $ O\left(\mathcal{G}\right)$.
As shown in \tabref{tabs1}, the latency of our \sysname is slightly higher than that of AutoBot.
With a larger $L$, the latency also marginally increases.
\begin{table}[!t]
\setlength{\tabcolsep}{7.5pt}
\renewcommand{\thetable}{S1}
\centering
\scalebox{1}{
    \begin{tabular}{c|cccc}
        \specialrule{\heavyrulewidth}{0pt}{0pt}
        Method  & AutoBot & $L=2$  & $L=4$
        & $L=8$ 
        \\
        \specialrule{\heavyrulewidth}{0pt}{0pt}
        Latency (ms)  & 7.72 & 8.05 & 8.12
        & 8.24 \\
        
        \specialrule{\heavyrulewidth}{0pt}{0pt}
    \end{tabular}
}
\caption{Latency comparisons between AutoBot and our \sysname at different granularities.
We measure the latency on one RTX 3090Ti GPU (batch size = 1), averaged over 1000 runs.
}
\label{tabs1}
\end{table}
\subsection{Sensitivity Analysis of Hyper-parameters}\label{hy}
\begin{figure}[!t]
\renewcommand{\thefigure}{S2}
\centering
\includegraphics[width=\linewidth]{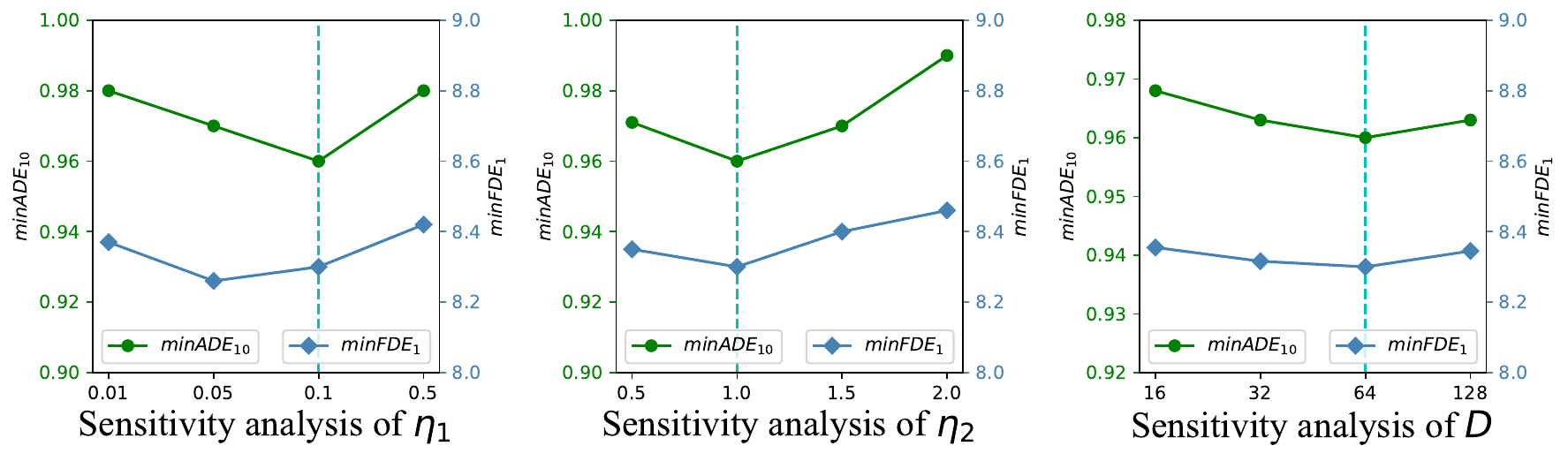}
\caption{Ablation studies for hyper-parameters. The cyan dotted lines denote the optimum parameters.
}
\label{fhp}
\end{figure}

We perform a sensitivity analysis on the hyperparameters, including $\eta_1$ and $\eta_2$ in Eq. (10), and the position embedding dimension $D$ in Eq. (5).
\figref{fhp} illustrates that our \sysname exhibits relatively low sensitivity to these parameters,  demonstrating its robustness across a range of values.

\begin{table}[!t]
\renewcommand{\thetable}{S2}
\setlength{\tabcolsep}{2pt}
\renewcommand{\arraystretch}{1.2}
\centering
\scalebox{0.85}{
    \begin{tabular}{c|ccccc}
        \specialrule{\heavyrulewidth}{0pt}{0pt}
        Model  & minADE$_5$$\downarrow$ & minADE$_{10}$$\downarrow$  & MR$_5$$\downarrow$
        & MR$_{10}$$\downarrow$ & minFDE$_1$$\downarrow$
        \\
        \specialrule{\heavyrulewidth}{0pt}{0pt}
        w/o PE  & 1.41 & 0.98 & 0.63 & 0.41
        & 8.35  \\
        $\mathcal{D}_p$  & \textbf{1.39} & 0.98 & \textbf{0.62} & 0.40
        & 8.33  \\
       $\mathcal{S}_p$ (ours)  & {1.40} & \textbf{0.96} & {0.63} & \textbf{0.39}
        & \textbf{8.30}\\
        \specialrule{\heavyrulewidth}{0pt}{0pt}
        $W_h=W_t$  & 1.42 & 1.00 & 0.64 & 0.41
        & {8.37} \\
        $W_h \neq W_t$ (ours)  & \textbf{1.40} & \textbf{0.96} & \textbf{0.63} & \textbf{0.39}
        & \textbf{8.30}\\
        \specialrule{\heavyrulewidth}{0pt}{0pt}
    \end{tabular}
}
\caption{Analysis of local recursive generation.
w/o PE indicates removing position embeddings.
$\mathcal{D}_p$ denotes the dynamic learnable embeddings, while $\mathcal{S}_p$ denotes the static embeddings \protect\cite{vaswani2017attention}.
$W_h=W_t$ denotes we use the same weight matrix for $W_h$ and $W_t$, while $W_h \neq W_t$ denotes we use the different weight matrices in Eq. (6).
}
\label{tabs5}
\end{table}

\subsection{Analysis of Local Recursive Generation}\label{lgp}
In this part, we analyze the sensitivity of position embeddings and weight matrices in Eq. (6) during the local recursive generation process. 
As shown in \tabref{tabs5},  
either $\mathcal{D}_p$ or $\mathcal{S}_p$ contributes to performance improvements.
Notably, the performance differences between $\mathcal{D}_p$ and $\mathcal{S}_p$ are minimal. 
Meanwhile, the performance gains achieved through position embeddings are marginal, indicating that the model obtains limited temporal information.
Additionally, employing distinct weight matrices for feature projection leads to a slight performance improvement. 
This can be attributed to the ability of these matrices to project the head and tail steps onto the same feature space, considering their value size.
\textit{\textbf{Notably, 
our main idea revolves around the global-to-local generation process, which sets it apart from the simultaneous and recursive paradigms. 
During this process,  
as shown in Eq. (6), we propose a simple yet effective approach for achieving local recursive generation.  
Alternative generation methods like CVAE and diffusion models can also be employed.
Generally, for these methods, it is crucial to prioritize both accuracy and efficiency.}}

\subsection{Analysis of The Regression Loss}\label{rl}
\begin{table}[!t]
\renewcommand{\thetable}{S3}
\setlength{\tabcolsep}{2pt}
\renewcommand{\arraystretch}{1.2}
\centering
\scalebox{0.9}{
    \begin{tabular}{c|ccccc}
        \specialrule{\heavyrulewidth}{0pt}{0pt}
        Model  & minADE$_5$$\downarrow$ & minADE$_{10}$$\downarrow$  & MR$_5$$\downarrow$
        & MR$_{10}$$\downarrow$ & minFDE$_1$$\downarrow$
        \\
        \specialrule{\heavyrulewidth}{0pt}{0pt}
    MAE  & 1.41 & 1.01 & 0.64 & 0.41
        & 8.42  \\
        NLL  & \textbf{1.38} & 0.99 & \textbf{0.63} & 0.40
        & {8.36} \\
        MSE (ours)  & {1.40} & \textbf{0.96} & \textbf{0.63} & \textbf{0.39}
        & \textbf{8.30}\\
        \specialrule{\heavyrulewidth}{0pt}{0pt}
    \end{tabular}
}
\caption{Analysis of the regression loss.
MAE indicates the Mean Absolute Error loss.
MSE denotes the Mean Square Error loss.
NLL denotes the Negative Log-likelihood loss.
Notably, when using NLL, we utilize the Gaussian model to describe the predicted distribution.
Subsequently, we utilize NLL to maximize the likelihood of the ground truths.
}
\label{tabs3}
\end{table}
In this section, we investigate the sensitivity of the regression loss in Eq. (4) and Eq. (9).
As shown in \tabref{tabs3}, 
our \sysname demonstrates robustness to these losses.
These loss functions all help enforce spatial constraints among key steps. 
Specifically, MSE contributes the most to performance gains, while MAE performs the poorest.

\subsection{Kalman Filtering for Post-processing}\label{kf}
\begin{table}[!t]
\renewcommand{\thetable}{S4}
\setlength{\tabcolsep}{1pt}
\renewcommand{\arraystretch}{1.2}
\centering
\scalebox{0.85}{
    \begin{tabular}{c|ccccc}
        \specialrule{\heavyrulewidth}{0pt}{0pt}
        Model  & minADE$_5$$\downarrow$ & minADE$_{10}$$\downarrow$  & MR$_5$$\downarrow$
        & MR$_{10}$$\downarrow$ & minFDE$_1$$\downarrow$
        \\
        \specialrule{\heavyrulewidth}{0pt}{0pt}
    AutoBot + KF  & 1.45 & 1.09 & 0.69 & 0.45
        & 8.73  \\
        ours  & {1.40} & \textbf{0.96} & \textbf{0.63} & \textbf{0.39}
        & \textbf{8.30}\\
        \specialrule{\heavyrulewidth}{0pt}{0pt}
    \end{tabular}
}
\caption{Analysis of Kalman filter during postprocessing.
KF assumes that the agents generally follow the constant velocity model as the dynamic model in Kalman filter \protect\cite{makansi2021exposing}.
}
\label{tabs6}
\end{table}
Given that the Kalman filter can be utilized for denoising and smoothing, it raises the question of whether it can handle postprocessing kinematically infeasible predictions to generate consistent trajectories. 
Employing the Kalman filter requires knowledge of the dynamic motion model for agents. 
However, it is difficult to determine whether these agents will proceed straight, turn left, or turn right.
Thus, as shown in \tabref{tabs6}, without the precise dynamic model for each agent,
Kalman filter fails to achieve superior performance.
\subsection{Visualization of The Selectable Granularity Strategy}\label{sgs}
\begin{figure}[!t]
\renewcommand{\thefigure}{S3}
\centering
\includegraphics[width=\linewidth]{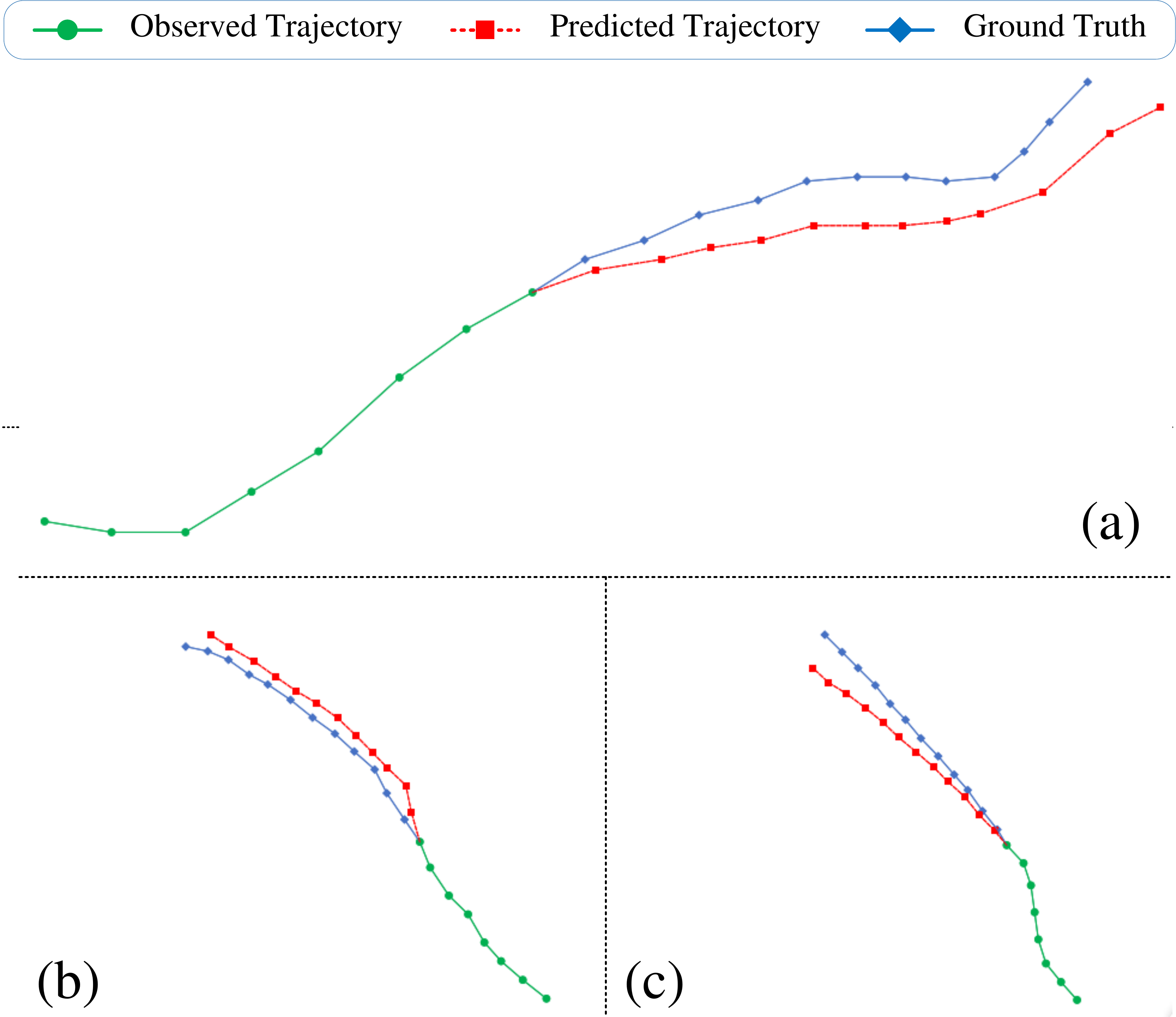}
\caption{We enumerate three cases 
in the UCY validation set, and visualize their predicted most likely trajectories.
The most likely trajectories for (a), (b) and (c) are generated with granularities of 2, 4, and 8, respectively.
}
\label{sm}
\end{figure}%
As shown in \figref{sm}, the optimal granularities vary across different agents. 
In particular, for \figref{sm} (a), finer granularity yields optimal trajectories by effectively capturing sudden motion changes.
Conversely, for \figref{sm} (b) and (c), 
coarser granularities generate optimal trajectories, as the model incorporates temporal constraints from more distant steps.

\end{document}